\documentclass[12pt,conference]{IEEEtran}
\IEEEoverridecommandlockouts
\usepackage{cite}
\usepackage{amsmath,amssymb,amsfonts}
\usepackage{algorithmic}
\usepackage{graphicx}
\usepackage{textcomp}
\usepackage{xcolor}
\usepackage{booktabs}
\usepackage{hhline}

\def\BibTeX{{\rm B\kern-.05em{\sc i\kern-.025em b}\kern-.08em
    T\kern-.1667em\lower.7ex\hbox{E}\kern-.125emX}}
\begin{document}

\title{Few-Shot Learning for Image Classification of Common Flora\\

}

\author{\IEEEauthorblockN{ Joshua Ball}
\IEEEauthorblockA{\textit{Edward E. Whitacre Jr. College of Engineering} \\
\textit{Department of Computer Science}\\
IEEE $\#94164957$ \\
Lubbock, USA \\
joshua.ball@ttu.edu}

}
\maketitle
\begin{abstract}
The use of meta-learning and transfer learning in the task of few-shot image classification is a well researched area with many papers showcasing the advantages of transfer learning over meta-learning in cases where data is plentiful and there is no major limitations to computational resources. In this paper we will showcase our experimental results from testing various state-of-the-art transfer learning weights and architectures versus similar state-of-the-art works in the meta-learning field for image classification utilizing Model-Agnostic Meta Learning (MAML). Our results show that both practices provide adequate performance when the dataset is sufficiently large, but that they both also struggle when data sparsity is introduced to maintain sufficient performance. This problem is moderately reduced with the use of image augmentation and the fine-tuning of hyperparameters. In this paper we will discuss: (1) our process of developing a robust multi-class convolutional neural network (CNN) for the task of few-shot image classification, (2) demonstrate that transfer learning is the superior method of helping create an image classification model when the dataset is large and (3) that MAML outperforms transfer learning in the case where data is very limited. The code is available here: github.com/JBall1/Few-Shot-Limited-Data
\end{abstract}
\section{Introduction}
The act of teaching some agent to learn a task in the field of machine learning is not a new one, however a great interest has grown in recent years over teaching an agent about learning how to learn. We want our agents to be able to mimic the human brains ability to quickly and efficiently learn.
In this case, we are speaking about meta-learning. Meta-learning performs outstandingly well in tasks of optimization, such as is showcased in the MAML paper where one task is to perform regression on data given by a sinusoidal distribution with parameters $p(\theta)$ [1]. Here, the goal is to perform well enough to achieve high accuracy results on data that is coming from that same distribution of $p(\theta)$ [1]. It is important to note that in meta-learning, unlike in transfer learning, the parameters for the optimization task are not trained at all. Meta learning is about the process of systematically learning to learn from how some $n$ of approaches work on some task $T$. Over time, the meta-learning algorithm will optimize its approach to the most-best method and \textit{learn} which is the most efficient; the algorithm does this by learning from its past iterations, commonly called meta-data, to learn other more complex o simple tasks faster.

Machine learning has been able to accomplish some incredible feats with image classification, whether it is for X-ray images of lungs for the classification of COVID-19 or to simply understand what you are eating. The same technology can be applied thanks to the incredible compute ability we know process in the modern world to simple tasks like identify flowers, no matter how rare or common they are. Our goal is to find some function $F$ which will be optimized for our task $T$ on our needed solution $S$. The task at hand here is the challenging problem of few-shot classification of five different common flora typically found around the world. 

We chose image classification as our most best solution over other possible solutions such as object detection, instance segmentation or semantics segmentation for several reasons. First, we do not desire to have any form of localization from our solution. Localization is important when you don't want to treat the image as a single object, and instead want to perhaps find several objects in that single image.
This is why the task of few-shot image classification is so important: there simply are so many use cases where the data is incredibly sparse.

One might draw the conclusion that if data is so sparse, why not attempt to do single-shot image classification instead of few-shot image classification. One-shot image classification is very difficult to achieve in the wild due to all the risks associated with it. For instance, that single image you have which you want to train an agent on $T$ to obtain some solution $S$ has a great likely hood of simply overtraining. The most difficult thing here is that you don't actually know if it over fit or not because you have a sample size of one. Another major issue with one-shot image classification is that the single image you have may not actually be representative of your $T$; in this scenario you will end up with a highly biased $S$ that will not accomplish what it was intended to do. 

Due to the many limitations that still stand with one-shot image classification, this is why few-shot image classification provides a reasonable but acceptable challenge: you simply have more data. Few-shot also allows you to rapidly test and tune a model without the need for high computational resources. This is ideal for when you want to test out some hyperparameters without having to use a high performance computing center or wait a long time for training to complete to test your theory on your model. Few-shot allows for the quick acquisition of task-dependant knowledge and slow extraction of transferable knowledge [6]. 

Transfer learning is heavily utilized today in applications which are not constrained by model size limitations or computational power for inferencing. Transfer learning allows one to utilize a highly performant convolutional neural network (CNN), remove the bottom layers which contain the weights and various other information related to the labels this model was trained on, and keep the top layers of the model [3]. The top layers of the top generic network are exceptional at feature extraction, which saves the new developer a great deal of time and resources. In our case, we have opted to use MobileNetV2 as our base model. We have found that, in comparison to other model architectures readily available for use in transfer learning such as VGG19, MobileNetV2 provides the best mix of size-on-disk with accuracy and training time. The general process of transfer learning can be seen in Fig 1 where we take MobileNetV2, $A$, remove the bottom layers and move $A'$ to the top of our new network.      
\begin{figure}[htpb]
    \centering
    \includegraphics[width=5cm, height=4cm]{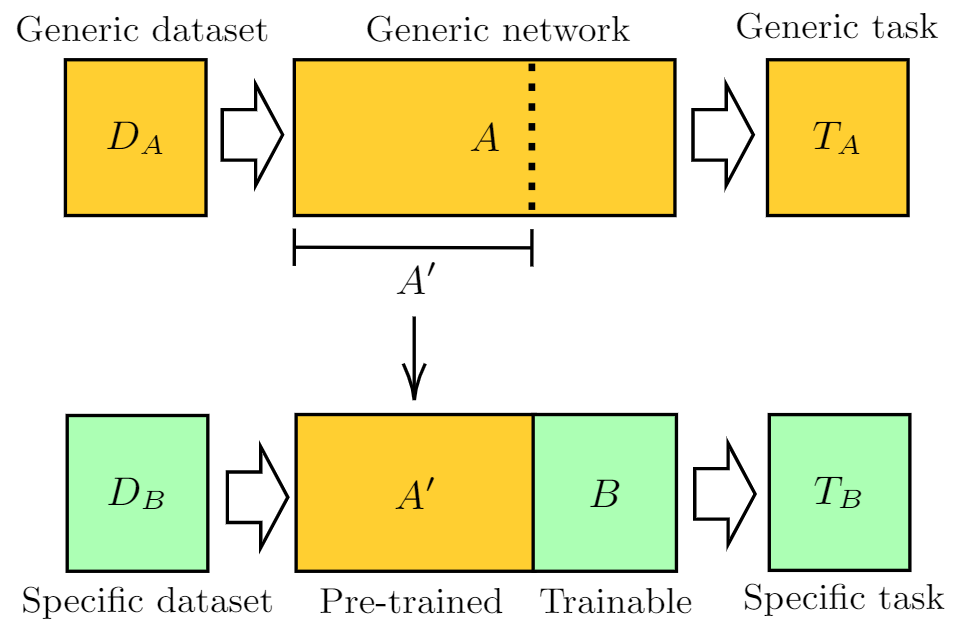}
    \caption{This graphic depicts the typical transfer learning scenario which is performed in this paper. }
    \label{Fig 1}
\end{figure}
\section{Related Works}
\subsection{Model-Agnostic Meta-Learning for Fast Adaptation of Deep Networks}
The authors propose an algorithm for meta-learning that, in comparison to other approaches, is compatible with any model trained with gradient descent [1]. They then showcase the results of their algorithm in several different experiments of various domains showing that, indeed, their approach of training on a small number of data points with a very small, set, number of gradient steps will and can produce good generalization on some task, $T$ [1].

Their algorithm, named model-agnostic meta-learning (MAML), is proven to show that it can find good initial model parameters for a variety of $T$, and with those parameters it can produce good results [1]. They essentially accomplish this by optimizing the initial parameters of their model and hope that it can adapt to the new $T$. This process is depicted in the figure below in Fig 2 [1].
\begin{figure}[htbp]
    \centering
    \includegraphics[width=5cm, height=4cm]{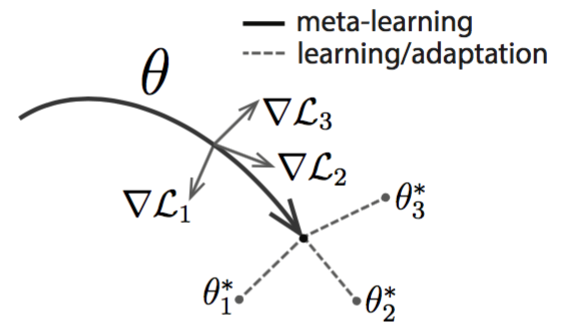}
    \caption{Here we see how MAML chooses its next $\theta$ for its next learning and adaption step in gradient descent. }
    \label{Fig 2}
\end{figure}
The process seen in Fig 2 was quite revolutionary at the time because it meant that there was a level of flexibility with meta-learning that was not yet available or seen with its methodology. Furthermore, this allowed for the possibility of generalization on a wide variety of tasks from simple linear models to complex tasks such as image classification.

\begin{figure}[htp]
    \centering
    \includegraphics[width=8cm, height=4cm]{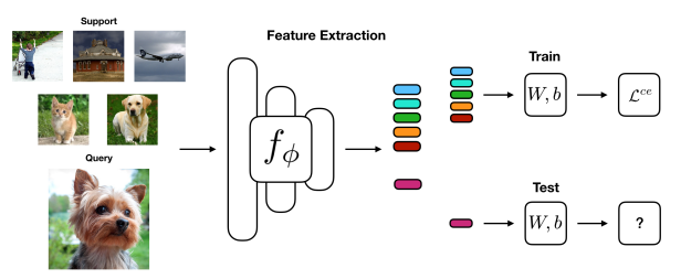}
    \caption{Here we same a graphical depiction of a 5-way 1-shot task: 5 support images and 1 query image are transformed into embeddings using the fixed neural network then fed into a linear model (logistic regression (LR) in this case). It is then trained on 5 support embeddings where the query image is tested using the linear model. The Output of parent through a softmax layer [1].}
    \label{Fig 1}
\end{figure}

In practice, MAML is more complex to setup than something like transfer learning, but the model performs extremely well when tasked with the generalization of data on a small number of classes that are fairly similar to each other. The paper showed that with Omniglot (Lake et ak., 2011) Model-Agnostic  Meta-Learning was capable of achieving a $98.7\%$ accuracy with 1-shot, 5-way accuracy learning and under 5-shot, 5-way accuracy learning was able to achieve a $99.9\%$ accuracy [1].

The results seen the Model-Agnostic Meta-Learning paper are impressive and overall showcase that their meta-learning method based on learning easily adaptable model parameters through few, and rather large, gradient descent steps is capable of creating a model that is able to generalize well on a wide range of tasks [1]. Furthermore, they show that their implementation can be adapted with nearly any model representation that is amendable to gradient-based training regardless of whether the task is classification, regression or even reinforcement learning [1]. Such demonstration is further explored in [2] where an adaptation of MAML is showcased with a linear classifier trained on the output of the student mode, capable of beating MAML's baseline performance numbers on large datasets like miniImagenet (Ravi $\&$ Larochelle, 2017).
\subsection{Few-shot Image Classification: Just Use a Library of Pre-trained Feature Extractors and a Simple Classifier}
The authors in this paper propose that using a series of high-quality, pre-trainined feature extractors for few-shot image classification along with a simple classifier on top will outperform any complex meta-learning algorithm [4]. The authors then showcase experimentally that using a library of pre-trainined feature extractors along with a simple forward-feeding network solve few-shot image classification well with a far simpler design and implementation than a state-of-the-art equivalent meta-learning algorithm such as MAML [4]. On average, the authors show that they beat MAML by as much as $18\%$ on the VGG Flower dataset [4].
\subsection{Rethinking Few-Shot Image Classification: a Good Embedding Is All You Need?}
In this paper the authors demonstrate that they are also able to beat out state-of-the-art meta-learning methods for few-shot image classification by first learning a supervised representation of the meta-training set, and then training a simple linear classifier on top of this meta-learning representation [2]. They also show that by utilizing the process of self-distillation they are able to boost their performance by as much $3\%$ further. The embedding model is carried over from the meta-training to the meta-testing and kept unchanged [1]. Unlike most approaches to few-shot, tuning is performed on the embedding model only [1].

Knowledge distillation has a teach model and a student model, where the student model is trained to the match the softened output of the teacher model, seen in Fig 3. The teacher model is pre-trained and therefore only backpropagation will occur on the student model. This differs grealy from transfer learning where the model weights are transferred from the pre-trained model to the new model. In transfer learning, the model complexity stays the same and often increases as well as the developer adds more layers [1].
\begin{figure}[htbp]
    \centering
    \includegraphics[width=5cm, height=4cm]{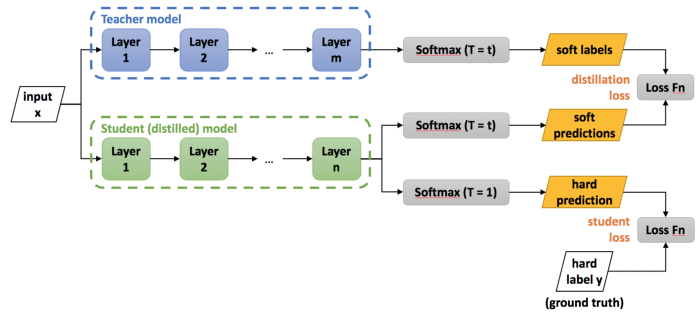}
    \caption{This graphic depicts parent-student model relationship and shows the self-distillation process which helps to compress the model with minimal loss. This then simplifies the output of our student model and allows a linear classifier to work well on its output. }
    \label{Fig 3}
\end{figure}
Learning a supervised or self-supervised representation of the meta-training set, followed by training a linear classifier on top of this representation, with something such as linear regression, is capable of outperforming many of the other state-of-the-art few-shot image classification methods such as MAML[1]. This shows that training a good teacher with a great embedding can beat even the most sophisticated meta-learning algorithms like MAML. The downfall of this is that even the approach suggested in this paper is significantly more complex than that of transfer learning. 
\subsection{Related Works Discussion}
From several related works it can be concluded that transfer learning indeed has some major advantages over meta-learning: (1) simpler to implement and (2) higher performance in the task of image classification and few-shot image classification. However, meta-learning for few-shot classification is still widely used in many applications due to its great generalization 'learning to learn' ability. These meta-learning algorithms, like MAML, are able to be trained very quickly and need significantly less computational resources than traditional convolutional neural networks built for image classification need. 

Each paper provides some interesting insight into which possible parameters and algorithms will yield the best result, but they also provide some ambiguity as well. For instance, with the MAML algorithm in [1], we can see that MAML nets an accuracy of $98.7\%$ in 1-shot learning with $K$ set to five for 5-way accuracy [1]. This result, however, was on the Omniglot dataset which is handwriting dataset [1]. In the same paper we see that MAML only proceeds to get a $48.7\%$ accuracy in 1-shot, 5-way accuracy [1] on the MiniImagenet dataset. This is worrisome because MiniImagenet is far more varied in data and image types compared to Omniglot, so one would assume it would in fact perform better. From this, we can start to formulate a strong hypothesis that perhaps MAML is not as robust in more complex tasks where embeddings and overall feature sets are very similar to each other. Regardless of all of this, accuracy on complex datasets is still significantly lower than that of traditional, data heavy, image classification. We do not believe that one could reasonably implement a few-shot algorithm into a high risk deployment where accuracy is more important than training time and model size. However, if your data is quite small we have found and will show later that small datasets perform better with MAML than with transfer learning. 

\section{Data and Processing}
The dataset utilized in this paper is based off of the Flowers Recognition dataset found on kaggle. \footnote{The Flower Recognition dataset can be accessed here: https://www.kaggle.com/alxmamaev/flowers-recognition} This dataset consists of 5 different classes on common flowers, or flora, that is commonly found around the world: dasiy, dandelion, rose, sunflower, tulip. Although there are potentially dozens of subspecies of each of these flowers, we have decided to keep it general for the sake of experimentation. 

\begin{figure}[htbp]
    \centering
    \includegraphics[width=8cm, height=2cm]{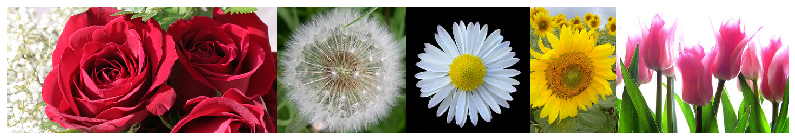}
    \caption{This graphic depicts one image from each of the 5 different classes. From left to right: Rose, Dandelion, Daisy, Sunflower and Tulip. }
    \label{Fig 4}
\end{figure}

We can see from Fig 3 that each class, from a glance, does look quite different from each other. This information is crucial for something like few-shot learning when data is already so sparse. We can also see that the images contain a wide range of colors. In some situations, transferring the images to greyscale is ideal, but in this situation where images could reasonably be confused in greyscale, such as with the sunflower and dasiy seen in Fig 4, it is best to keep our dataset in RGB.

\begin{figure}[htbp]
    \centering
    \includegraphics[width=5cm, height=4cm]{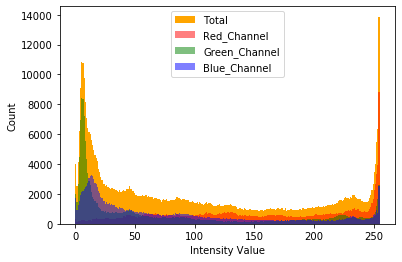}
    \caption{This graphic depicts the histogram plot of the 3 different RGB color spectrum for the rose image seen in Fig 3. }
    \label{Fig 5}
\end{figure}

Our data split consists of two different versions, first the complete set for testing and comparison purposes and then a subset of the entire set. The complete set consists of the 5 different classes with around 1000 PNG color images per class with a $70\%$ split for training, $25\%$ for testing and $5\%$ validation. The subset consists of 25 images for training, 5 for testing and 25 for validation. The images were split randomly to ensure the greatest diversity of features and to minimize bias. 
\subsection{Data Augmentation}
Given our small sample size for experimentation purposes, the use of data augmentation is a critical tool to increase data size without having to somehow get more images. Image augmentation allows for the artificial inflation of data by applying specified image augments to your training set. This process occurs on demand as the model is training, not requiring one to manage several dataset versions or transfer large datasets which is very handy. An important note is to set a seed value, otherwise your ability to reproduce your results will be limited. In table I the various image augments and their respective values can be seen.
\begin{figure}[htp]
    \centering
    \includegraphics[width=5cm, height=4cm]{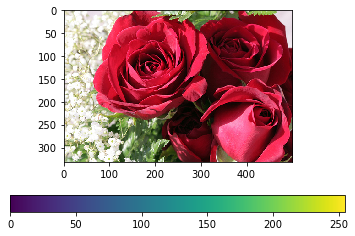}
    \caption{Here we can further see the color distribution of the same image explored in Fig 5.}
    \label{Fig 6}
\end{figure}
Image augments applied to our dataset allowed us to increase the variance in our dataset, ideal for training a model like a CNN. These augments were also applied to the test and validation sets through an image data generator from the Keras library. These augments effectively generate 10 additional images for every image originally in our dataset. The augmentations are capable of generating images from a single image by applying several different changes to the image randomly, such as: rescaling the image, rotating the image some number of degrees on the $x$ axis, shifting the image some percentage on the $x$ axis, shifting the image some percentage on the $y$ axis, shearing the image, flipping the image entirely over horizontally, and also by filling the modified image into our predefined 224 by 224 pixel range. The particular value used for these various augments can been seen in Table I.

\begin{table}[htb]
\centering
\begin{tabular}{cc}
\toprule
\textbf{Parameter}            & \textbf{Value} \\ \bottomrule
Rescale            & $1/255$\\ 
Rotation range     & 40                 \\
Width shift range  & 0.2                \\
Height shift range & 0.2                \\
Shear range        & 0.2                \\
Zoom range         & 0.2                \\
Horizontal flip    & True               \\
Fill mode          & Nearest            \\ \bottomrule

\end{tabular}
\caption{Image augments applied to training sets. [3].}
\label{Table 1.}
\end{table}

Image data augments here are used to extend the size of our training dataset in order to better improve the performance the performance of our models. Our performance improves as we reduce the variable of data sparsity, allowing our model to generalize better. The images in our actual dataset are not fed to the model when utilizing image augmentation. Instead, only the augmented images are provided to the model during training.

\section{Method}
We will first establish baseline performance standards for meta-learning and transfer learning. We will then present our baseline performance and tune each respective algorithm until we achieve the best performance possible for each. For MAML, we train on an image classification task on the merged meta-training data to learn an embedding for our $T$. The model trained here is then used at meta-testing time to extract embedding for a simple linear classifier. This works because the features extracted at this point are very simple, as a linear classifier would not work for $T$ where we are classifying five different classes of images.

\section{Experiments}
Our experiments will revolve solely around the task of image classification on the flower dataset. We then validate our experiments by testing the model on the complete flower set's test set. From this testing, we generate the images seen in Fig 7 and Fig 8.
\subsection{Setup}
Our setup will consist of three different models being tested: (1) traditional transfer learning model with the complete flower dataset which will be used as our baseline for performance, (2) an adaption of our model in (1) which will be capable of converging on a dataset as small as our subset and (3) our tuned MAML implementation which will be trained on our flower data subset. Each model will be trained sufficiently until early stopping and until it can be assumed that no further level of tuning would reasonably improve the performance of the model. 

\subsection{Model-Agnostic Meta-Learning model setup and tuning}
For the implementation of Model-Agnostic Meta-Learning, we utilized a Pytorch adaption of the original implementation seen in [1]. This implementation is nearly identical to the original papers implementation but instead uses Pytorch which was found to be faster to converge and easier to modify for various datasets. The particular version of the implementation was a modified version of that seen in [7].

\begin{table}[htb]
\centering
\begin{tabular}{cc} 
\toprule
\textbf{Parameter}            & \textbf{Value}    \\ \bottomrule
Epochs               & 60000    \\ 
Image Color          & 3 - RGB  \\ 
Image Size           & 84x84    \\ 
K-Query/Split             & 15/1\\ 
Meta Learning Rate   & 0.001    \\ 
N-Way                & 5        \\ 
Task Number          & 4        \\ 
Update Learning Rate & 0.01     \\ 
Update Step          & 5        \\ 
Update Step Test     & 10       \\ \bottomrule
\end{tabular}
\caption{The MAML parameters used in our implementation.}
\end{table}

We utilized the initial parameters seen in Table II, which is mostly based off of the original implementation from [1]. We found these parameters to perform quite well, resulting in a training and testing accuracy of $100\%$ in nearly a quarter of the time it took to get significantly better results than our transfer learning with the subset. 

\subsection{Hyperparameter Tuning for Transfer Learning }
In order for us to get our Convolutional Neural Network to reach state-of-the-art performance, we had to tune the various hyperparameters available to us extensively. This process requires several different variants of the CNN to be generated and tested. Ultimately we decided on utilizing a linear sequential model, with MobileNetV2 added to the top of our model with the imageNet weights [3]. We then followed up out model global average pooling 2D layer which feeds directly into a ReLU activation layer with a dropout of 0.7, which is a very high dropout but helps avoid over training [3].

\begin{table}[htp]
\centering
\begin{tabular}{|c|c|c|c|c|}
\hline
\multicolumn{5}{|c|}{\textbf{CNN Variation 2}}                            \\ \hline
Learning Rate & Optimizer & Dropout & L2  & Accuracy    \\ \hline
0.0001        & Adam      & 0.7      & 2e-6           & 0.965   \\ \hline
\multicolumn{5}{|c|}{\textbf{CNN Variation 5}}                           \\ \hline
0.001          & Adam      & 0.5     & 2e-6           & 0.932    \\ \hline
\end{tabular}
\caption{We tried 7 different configurations but went with configuration 2.}
\end{table}

\section{Interpreting our Results from Experiments}

In order to properly understand and interpret our models results, we must first explore a few different topics first. The F-score, or F1-score is a measure of a model's accuracy on a dataset [8]. The F-Score is a mathematical formula which is a combination of the precision and recall of the model, and is defined as the harmonic mean of the model's precision and recall [8]. The F-Score is usually used for the evaluation of retrieval systems such as search images, and for most types of machine learning models such as ours for image classification [8]. It is possible to adjust and skew the F1-Score depending on your application and operator needs, however. This adjustment is done by refining the importance of precision and recall [8]. The formula for the standard F1-score is the harmonic mean of the precision and recall [8]. A perfect model would have a F1-score of 1 [8].

F-score is broken down into several different variables, seen below. Precision is the fraction of true positive examples that the model has classified as positive, or in our case the proper label like rose [8]. Recall is related to the models sensitivity [8]. Recall is represented as the fraction of examples classified as positive among the total number of positive examples [8]. True positives, or $tp$, represent the number of true positives classified by the model seen in the formula below. False negatives, or $fn$, is the number of false negatives classified by the model. Finally, we have false positives, or $fp$, which are false positives classified by the model.

\begin{figure}[htp]
    \centering
    \includegraphics[width=8cm, height=1cm]{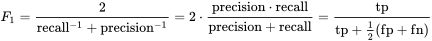}
    \caption{The F1-Score Formula}
    \label{Fig 10}
\end{figure}

All of these, false positives, false negatives and true positives are important things to keep in mind when testing a model and are especially important depending on your models task. Depending on your task a $fp$ may end up being much more devastating than a $fn$ or $tp$ if it is incorrect. There is a certain degree of risk when taking into account this factor and this is why F1-Score is so important for someone who is developing a model to see and test. 

\subsection{Results with Transfer Learning on complete dataset}
Our results with transfer learning, seen in Table III, netted excellent generalization for our model. Our models performance was then tested on the test set and is depicted in Fig 7 on a confusion matrix. From the figure, we can see that the model performs very well, with only some issues with tulips and roses, which is expected due to the fact that we did not separate any of the flowers by their specific species, only by their genus. 

\begin{figure}
    \centering
    \includegraphics[width=8cm, height=8cm]{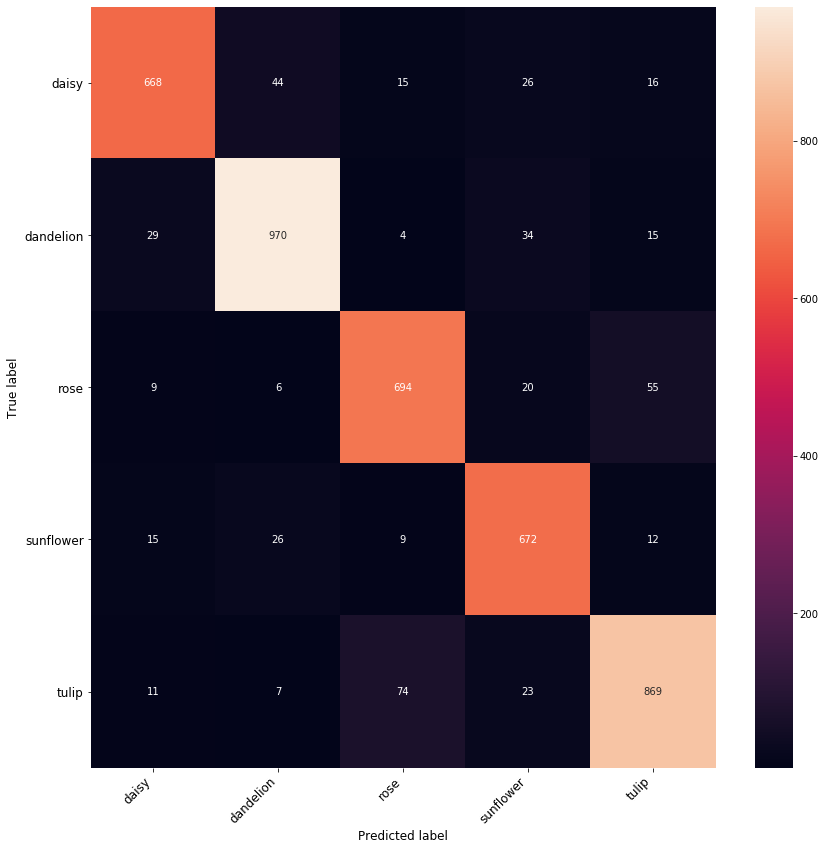}
    \caption{This graphic depicts the confusion matrix of the performance of our transfer learning model, which was trained on the complete flower dataset.  }
    \label{Fig 7}
\end{figure}

The performance here again is trained on the complete flower dataset with over 4000 images, and will serve only as our baseline for transfer learning. These results represent what one might expect utilizing traditional image classification methods along with transfer learning which requite a great deal of data and computational power to complete in timely manner.

\begin{figure}[htp]
    \centering
    \includegraphics[width=4cm, height=3cm]{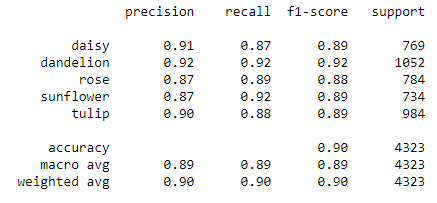}
    \caption{Our transfer learning models performance on the complete flower dataset. We see the models performance on various parameters such as recall and f1-score, overall with an impressive accuracy of $90\%$.  }
    \label{Fig 8}
\end{figure}

In Fig 10 we can see the various different performance values for our transfer learning model which was trained on the complete flowers dataset. We can see quite a robust response from our model, with impressive F1-scores for every class nearly hitting $90\%$. We can also see the recall scores around the same level as our precision scores, showing that the model is able to differentiate our false positives, true positives and false negatives quite well. Overall our model, on this particular test set, ended with a weighted average accuracy of $90\%$.

\subsection{Results with Transfer Learning with limited data}
In order to properly tune our model for the task of few-shot image classification, several tuning and optimizations had to be performed to help the model adapt to the lack of data. This process requires great attention to detail, as you must ensure the model does not overfit on the limited data, but also that the model diverges to a solution $S$. To do this, we utilized several technologies such as learning rate optimizers, dropout layers, early stopping, L2 regularization tuning, as well as trying several different optimizers. We tuned our L2 regularization, commonly known as ridge regression, to bring out weights to close-to zero value [3]. 

\begin{figure}[htp]
    \centering
    \includegraphics[width=4cm, height=3cm]{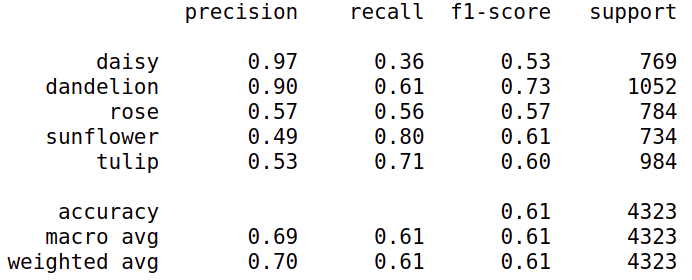}
    \caption{Our transfer learning models performance on the subset, such as recall and f1-score, overall with an accuracy of $61\%$.  }
    \label{Fig 8}
\end{figure}

From our models test scores, which were generated on the complete flower dataset's test set we can draw several observations. First, we can see that although the models precision is fairly high across the board, there is a distinct issue with precision on the rose and tulip classes. This is likely due to the fact that roses and tulips both have sub-species which look like each other. Additionally, roses and tulips both have representation in the dataset with similar colors such as red and pink and due to the data being so limited, it seems the model has focused on this particular feature more than it should have. In Fig 11 we can see the various different performance values for our transfer learning model, similarly to what we saw in Fig 10. As previously discussed, however, we can see this models responses was not nearly as robust as our complete dataset variant. Particularly, we can see that recall, or the models sensitivity, was very poor for classes of daisy, rose and dandelion. Recall is determined by taking the number of true positives, $tp$, and dividing them by the number of $tp$ plus false negatives, $fn$, such as seen below in equation (1):

\begin{equation}
    Recall = \frac{tp}{tp+fn}
\end{equation}

\begin{figure}[htpb]
    \centering
    \includegraphics[width=8cm, height=7cm]{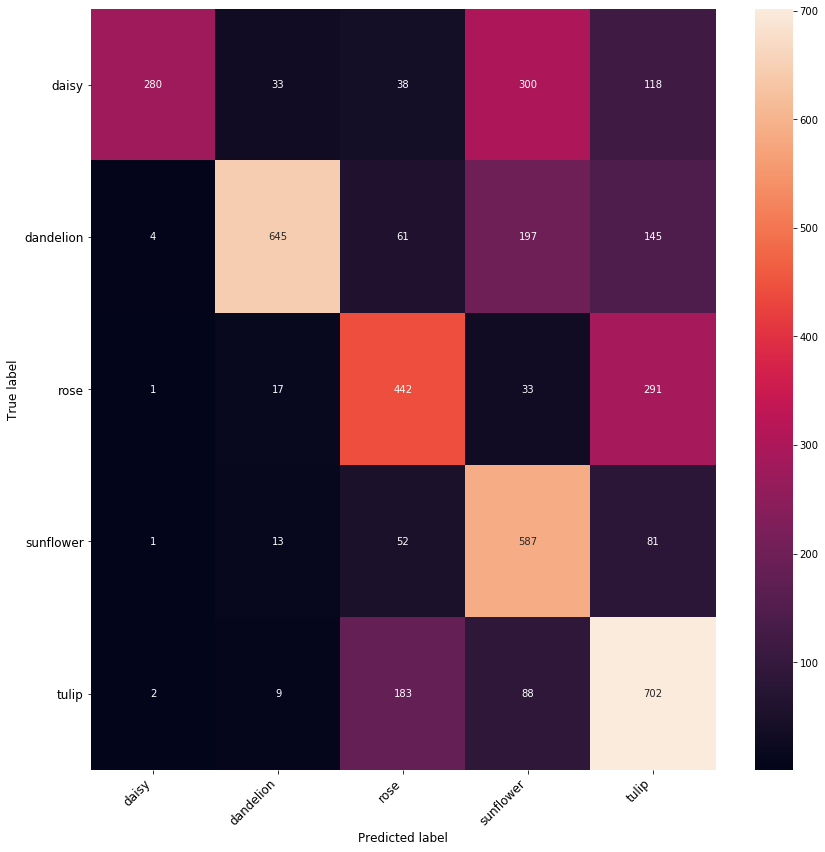}
    \caption{We see the confusion matrix for the transfer learning model trained on the limited subset of the flowers dataset. We can see that the model tends to confuse Sunflowers and Daisy's as well as Tulips and Roses the most. This can likely be attributed to the similar petal shapes and overall shape the flowers have.  }
    \label{Fig 9}
\end{figure}

Similarly to the confusion matrix seen in Fig 8, we see that the limited dataset model struggles greatly to generalize. This is a common fault of models that suffer from data sparsity and ours is no different. Despite our best efforts to continue to tune out model, an accuracy of $61\%$ was the best we could achieve.

\begin{figure}[htp]
    \centering
    \includegraphics[width=5cm, height=5cm]{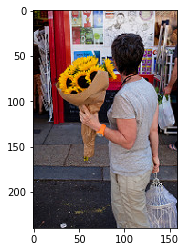}
    \caption{Here we see the depiction of an image which was incorrectly predicted with $97\%$ accuracy as a Rose from the complete test set. This image is a good example of how a neural network can struggle with generalization when it's training data has been so sparse.}
    \label{Fig 10}
\end{figure}

\subsection{Results with Meta-Learning}
Our results with utilizing Model-Agnostic Meta-Learning on the limited dataset were excellent. With MAML we ended up, through experimentation, with a testing and training accuracy of $100\%$ on our flower subset. This performance is of course excellent, but, as we can see from [1], MAML shines with small amounts of data like Omniglot (Lake et ak., 2011) where the features are not very different from each other. But from [1] we can also see that MAML struggles to touch the state-of-the-art performance seen with traditional transfer learning with large amounts of data. This is likely due to the model guessing incorrectly on its next large gradient step. In total, it took 990 steps to reach the performance we were happy with on MAML.  

Meta-Agnostic Meta-Learning is able to take gradient steps that demonstrate a certain level of intelligence. The algorithm is able to adjust its parameters efficiently and perhaps most importantly,  effectively enough to where it is capable of beating out our transfer learning model using the imageNet weights [3].

\begin{figure}[htp]
    \centering
    \includegraphics[width=6cm, height=6cm]{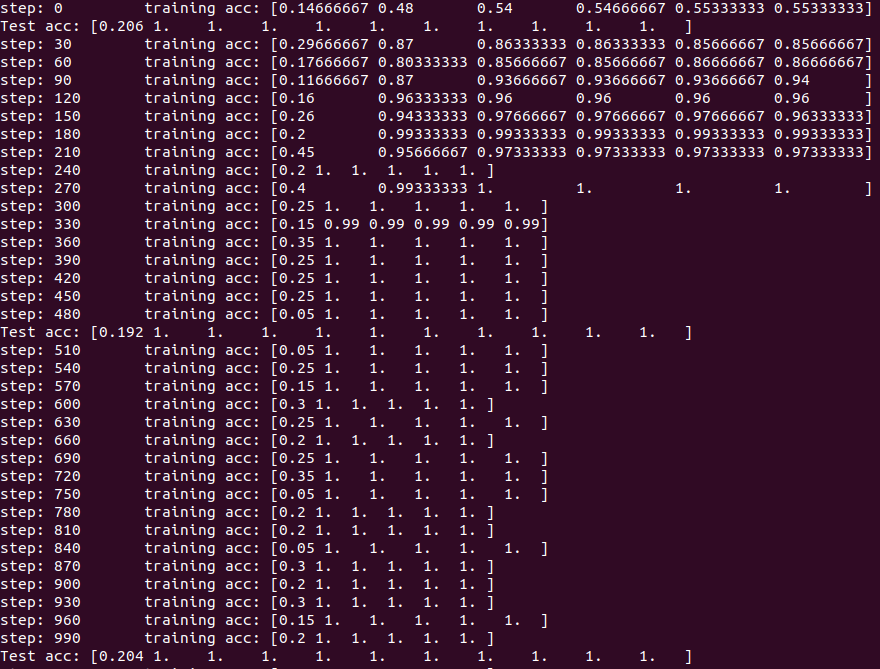}
    \caption{Here we can see an image from the training of the Model-Agnostic Meta-Learning model. This image shows several important things such as the training and test accuracy per gradient step. }
    \label{Fig 10}
\end{figure}

\section{Conclusion}
In this paper we explored the creation of several different classifiers for the task of image classification on five different types of flowers commonly found around the world. We developed a robust transfer learning model based off on MobileNetV2 and tuned it extensively to get state-of-the-art performance on our complete dataset with an accuracy of $96\%$ on the validation set; this model was then used as our baseline for performance to compare against our other two implementations. Our next implementation was very similar to our baseline model but it was tuned to ensure it would converge on our much smaller data subset it was trained on.
\section{Discussion and Future Work}
The results show us that Model-Agnostic Meta-Learning performs extremely well in the task of classification where data is very limited, beating our transfer learning implementation by $39\%$. Convolutional Neural networks have constantly been demonstrated as being successful in representing the connected ability we see in the human brain and they make use of the hierarchy-style patterns seen in our image dataset. This is why both MAML and our transfer learning implementation perform very well despite being starved of data. However, the power of MAML and its ability to create smaller, state-of-the-art models with less computational resources and time needed are extremely valuable.

We initially perceived that it may be the case that MAML would not be able to match transfer learning, even in the few-shot environment where it shines for out use case of flower classification but this was incorrect. As was demonstrated, MAML beat both implementations of our transfer learning model. It would be beneficial to test our MAML implementation on the complete dataset to see if it is capable of matching our transfer learning model on the complete dataset, however. As was seen in [1], MAML tends to struggle on datasets that have a large number of highly varying features, like those seen in imageNet. Furthermore, it would be beneficial to test several different backbones on the transfer learning model and continue to tune the MAML model by adding a linear classifier on top of it, as seen in [2]. These changes would likely greatly increase our models ability to generalize on larger and more complex datasets therefore increasing their use cases in more and more complex tasks.
\section*{Broader Impact}
We would like to help bring few-shot classification to the broader market of tasks where the data is very sparse and simply can not be increased. Some scenarios of this would include the classification of rare and deadly diseases. These diseases could help be identified through the use of image classification and object detection, even on the cellular level.

Few-shot image classification can also be used in our primary example in this paper: the classification of flowers. We find the classification of flowers to be particularly interesting due to the incredibly challenging task it is to do so. Flowers come in so many different colors, shapes and sizes which really bring out the challenge so much because two flowers could look nearly identical but have a few small features which separate them. Here, the task of the image classifier comes into play to find these features, no matter how small, and extract them and generate weights for these various labels. This situation may not be an issue with something like a common daisy or garden-variety rose where data is quite plentiful and images exist of nearly every possible angle. However, this is an major issue for potentially rare flowers, such as the Chocolate Cosmos which are thought to be extinct in the wild. If you are trying to classify or be able to classify one of these rare flowers, technology such as few-shot and perhaps even single-shot in extreme cases is needed to develop an even somewhat reliable and hopefully robust classifying agent. From our performance with the Model-Agnostic Meta-Learning algorithm, we can confidently say this technology is quite impressive and will work wonderfully for our task of image classification of common flora, or flowers.

Although Model-Agnostic Meta-Learning has been shown here, in [1] and [2] to be very successful at the task of few-shot image classification there lies but one single problem: potential bias. Bias is a major issue today in machine learning, and for good reason. Whether your data is skewed to unintentionally focus on a group of people or signals, it still brings up an inherent issue with your model and its reliability. Bias is a difficult issue to get around, especially with few-shot image classification where data is often very sparse, like in our case, and images may have to be hand picked. When data is hand chosen, a great deal of personal bias has been introduced by the operator by what they find important. The designers of these models, like us in this paper, can go as far as they can to ensure that bias is not introduced by picking images randomly to be added to each dataset and it's splits but even then the 'randomness' of how these images are picked double be biased towards a certain time or value such as even or odd numbers. It is the responsibility of the classifying agent's designer to ensure bias is avoid at every turn of the model and data creation process for the betterment of society and for the task they are working on.

\section*{Acknowledgment}
The author would like to thank Zohreh Safari for her helpful discussions and technical advice, Dr Victor Sheng for his direction, initial technical guidance and instruction on this topic.

\section*{Author Information}
\textbf{Joshua Ball}, Graduate Student, Department of Computer Science, Texas Tech University 

\vspace{12pt}

\end{document}